\documentclass[runningheads]{llncs}
\usepackage[T1]{fontenc}
\usepackage{graphicx}
\usepackage{caption}
\usepackage{subcaption}
\usepackage{blindtext}
\usepackage{comment}
\usepackage{floatrow}
\usepackage{multirow}
\usepackage{comment}
\usepackage{parskip}
\usepackage{enumitem}
\usepackage[dvipsnames]{xcolor}
\newcommand{\squeezeup}{\vspace{-2.5mm}}
\usepackage[misc,geometry]{ifsym}
\begin{document}
\title{A Self-Supervised Pose Estimation Approach\\ for Construction Machines%\thanks{Supported by DAAD.}
}

\author{Ala'a Alshubbak \inst{1,2}(\Letter) \and Daniel Görges\inst{1} }

\authorrunning{A. Alshubbak and D. Görges}
\institute{Institute of Electromobility, University of  Kaiserslautern-Landau, \\ Kaiserslautern, Germany \and German Jordanian University, Amman, Jordan.\\
\email{\{al-shubbak,goerges\}@eit.uni-kl.de}}
\maketitle              % typeset the header of the contribution

\begin{abstract}
Pose estimation is a computer vision task used to estimate a skeleton of dynamic systems to predict future movements. Most of the research in this direction is based on a supervised learning approach which requires a massive amount of labeled datasets. 
In this paper, a self-supervised three-stage model based on a contrastive learning approach is introduced for estimating a skeleton of dynamic construction machines; such as excavators without using any labeled images for the first stage. The whole model structure is divided into three stages: the pre-train stage using the SimCLR contrastive approach, and two fine-tuning stages for the transfer learning and the downstream task. The model can leverage the features and learn from a huge unlabeled dataset called ACID to two small datasets generated from NVIDIA Isaac and MATLAB Simscape simulators as well as transfer the knowledge to a smaller dataset with a ratio of 3.5\% from the original ACID dataset. The results show that the proposed approach can improve the accuracy of pose estimation for heavy construction machines in real images by 11\% and 13\% in comparison to the normal self-supervised approach with two backbones ResNet-50 and HRNet-W32, respectively.

%in 150--250 words.

\end{abstract}

\section{Introduction}
Pose estimation is a computer vision task that is used to estimate a skeleton of dynamic systems to predict future movements \cite{dang2019deep,sarafianos20163d}. Most of the research in this direction is based on a supervised learning approach which requires a massive amount of labeled data \cite{pereira2019fast,graving2019deepposekit}. 
To the best of the authors' knowledge, there is no benchmark data for pose estimation for construction machines produced yet. 
The only benchmark dataset in this field is the ACID dataset \cite{xiao2021development}, which consists of 10K images of 10 different classes of heavy construction vehicles, such as excavators, backhoes, and trucks. All of these are only labeled in the form of bounding boxes. % \textcolor{blue}{Figure \ref{fig4:sample from acid original} shows some samples from the dataset.}
This labeled format is only suitable to work with downstream tasks such as object detection and segmentation, which makes the task of pose estimation for such objects a challenging topic. 
%multi image one caption for only acid dataset 
%\begin{figure}[htb]
%    \centering
%    \begin{subfigure}[b]{0.80\textwidth}
%        \centering        \includegraphics[width=0.475\linewidth]{pictures/acid/bbox/001113.jpg}%
%        \hfill
%        \includegraphics[width=0.475\linewidth]{pictures/acid/bbox/000009.jpg}
%    \end{subfigure}
%    \vskip\baselineskip
%    \begin{subfigure}[b]{0.80\textwidth}
%        \centering
%        \includegraphics[width=0.475\linewidth]{pictures/acid/bbox/000238.jpg}%
%        \hfill
%        \includegraphics[width=0.475\linewidth]{pictures/acid/bbox/000179.jpg}
%    \end{subfigure}
%    \caption{Images samples from ACID dataset with bounding box annotation.}\label{fig4:sample %from acid original}
%\end{figure} 
%\squeezeup
In this paper, a self-supervised contrastive learning model is designed to estimate a skeleton of dynamic heavy machines without using any labeled images, as the model depends on data augmentation to learn the visual representations that distinguish objects from one another in different images. The model leverages the concept that images produced from the same source have similar visual representations and features, while the ones produced from different sources will have a variant representation. The model is trained on a huge dataset and then fine-tuned on smaller ones. By reusing the structure of the SimCLR model \cite{chen2020simple}, the proposed model can detect the keypoints of heavy machines and produce a skeleton in different situations and positions. The model outperforms the supervised approach on two simulated datasets from MATLAB Simscape Multibody \cite{Exca2023Simscape} and NVIDIA Omnivers environments by 1.7\% and 0.8\% AP, respectively. In addition, it is fine-tuned over a small amount of the dataset and gives an increase of 11\% and 13\% AP for two different backbone models. 

The contributions of this work are:
\begin{itemize}[noitemsep]
\item Automatically generate a synthetic dataset with annotation using simulators for pose estimation tasks. This data represents mainly one object class of excavator. 
\item Design and implement a self-supervised deep neural model based on the SimCLR model to generate a skeleton for a heavy construction machine ‘excavator’ using a smaller amount of annotated dataset. 
\item Minimize the domain shift between the simulated situation and the real situation using a small dataset from a real environment, which improves the model's performance on real-world data.
\end{itemize}

This paper is divided into five sections: Section 2 introduces the previous works, while Section 3 contains the problem statement and the model architecture proposed to solve such a problem. Section 4 explains the experiment setup and the datasets. The results and discussion in section 5 provide a detailed explanation of the performance of the proposed approach. Lastly, a conclusion and future direction are given in section 6.

\section{Related Work}
\subsection{Pose Estimation In Heavy Construction Machines}
The interest in safety when using heavy construction machines has grown rapidly in recent years, alongside the development of artificial intelligence models and deep learning. The advancements aim to enhance accuracy and reliability in detecting objects and predicting motion for both humans and machines. Researchers primarily focus on human safety by detecting, tracking, and predicting human motion. However, there is less emphasis on detecting and tracking the motion of construction machines themselves  \cite{pham2021current,luo2020real,lin2021temporal}.
One research study discussed a method called the Monocular Vision Marker System \cite{zhao2021pose}, which uses a monocular camera to detect markers on an excavator's manipulator parts. This system consists of two phases: camera calibration and pose estimation using the CALTag marker-based approach. The model achieved an error of fewer than 8.5 degrees in orientation and less than 22 millimeters in position.
Within the fewer attempts which have been made to develop pose estimation models for heavy machines using deep neural networks, two main methods were mentioned in the literature. The first method detects the essential keypoints that form the motions in the machines \cite{luo2020full,luo2021construction}, while the second method detects the machine parts and then extracts a skeleton \cite{soltani2017skeleton}.

 One study \cite{luo2020full} designed a pose estimation model using supervised learning with a deep neural network based on a Stacked Hourglass Network (HG) and a Cascaded Pyramid Network (CPN). The model achieves good results in terms of normalized errors, correct keypoints, and area under the curve with six keypoints. Another study \cite{luo2021construction} built a framework using a Gated Recurrent Unit (GRU) to predict the pose estimation for excavators and dump trucks, achieving an average accuracy of 90.22\% of correct keypoints (PCK). The whole dataset used within such a deep neural network consists of 53484 annotated frames.

\begin{table}[htp]
\begin{tabular}{cc}
\includegraphics[height=1.25in]{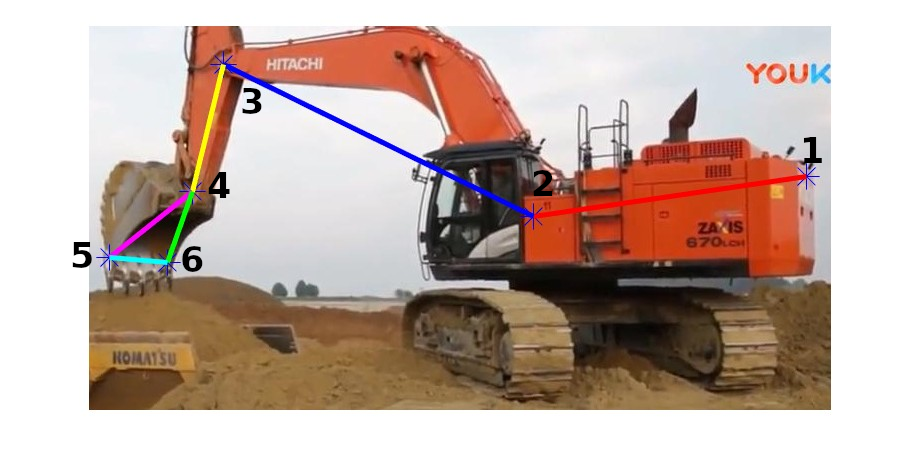} & \includegraphics[height=1.25in]{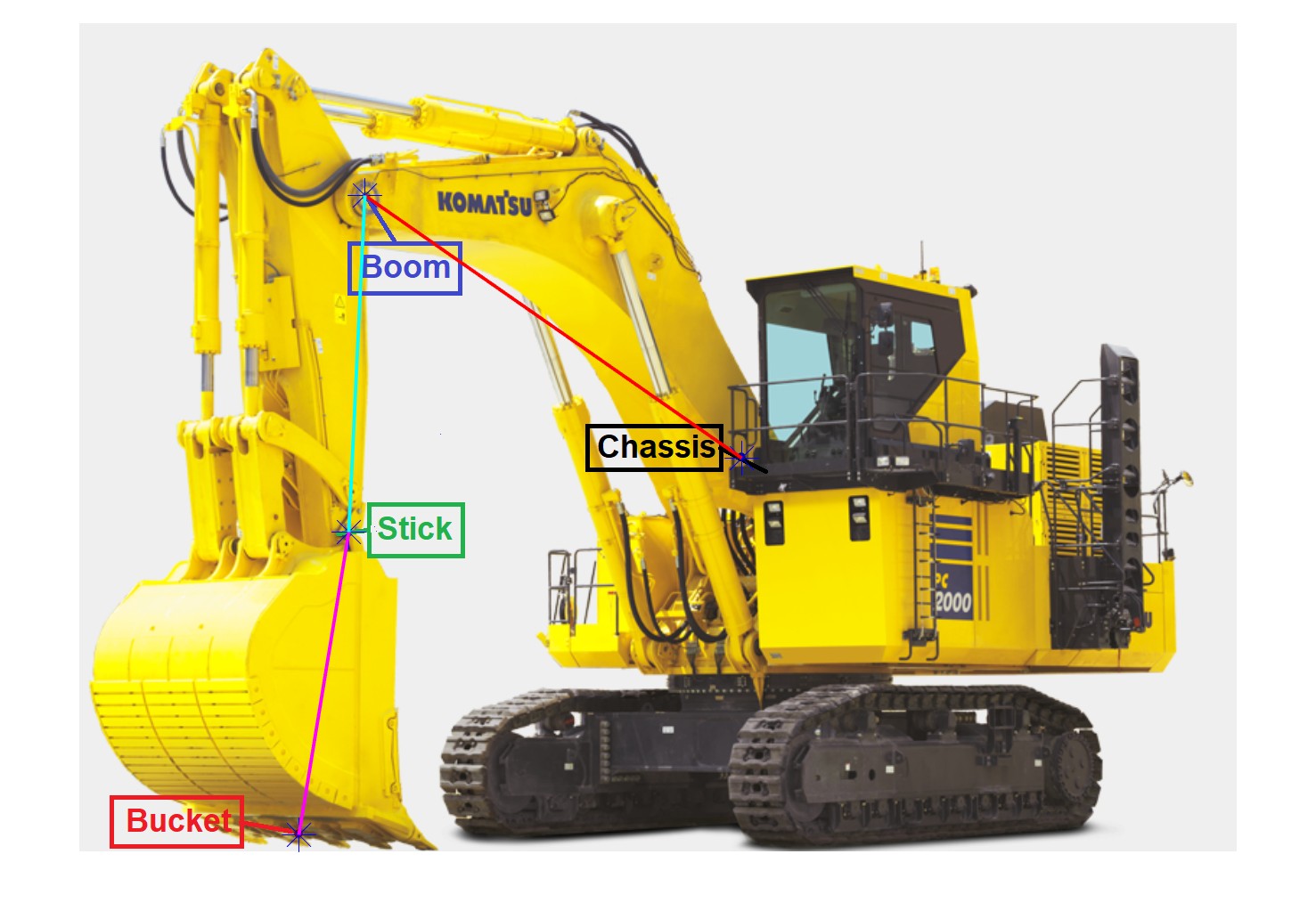}\\
\begin{tabular}{cc}
      6 Keypoints & Definition \\ \hline
        1 & "Body\textunderscore end" \\
        2 & "Cab\textunderscore boom" \\
        3 & "Boom\textunderscore arm" \\
        4 & "Arm\textunderscore bucket" \\
        5 & "Left\textunderscore bucket\textunderscore end" \\
        6 & "Right\textunderscore bucket\textunderscore end" \\
      \hline
      \end{tabular}  & 
       \begin{tabular}{cc}
     4 Keypoints & Definition \\ \hline
        1 & "Chassis" \\
        2 & "Boom" \\
        3 & "Stick" \\
        4 & "Bucket" \\
        \hline
      \end{tabular}\\
\end{tabular}
  \captionof{table}{Definition of machine keypoints in literature and this work.}\label{tab:table1}
\end{table}

In contrast, another research \cite{soltani2017skeleton} proposal used traditional machine learning methods to estimate the skeleton of excavators by detecting their parts. This approach involved multiple stages, including multiple detectors, a cluster model, and an orientation estimator. Although this method suffered from missing some data or keypoints, it achieved reasonable accuracies for detecting the stick, boom, and chassis parts of the excavator.
Most efforts in designing machine pose estimation models rely on a large dataset of annotated images, either synthetic or real, and training deep neural networks using a supervised learning approach.

\subsection{The State-of-Art: Self-Supervised Learning Approach}
\subsubsection{Self-supervised learning (SSL)} 
\begin{figure}
\includegraphics[width=0.75\textwidth]{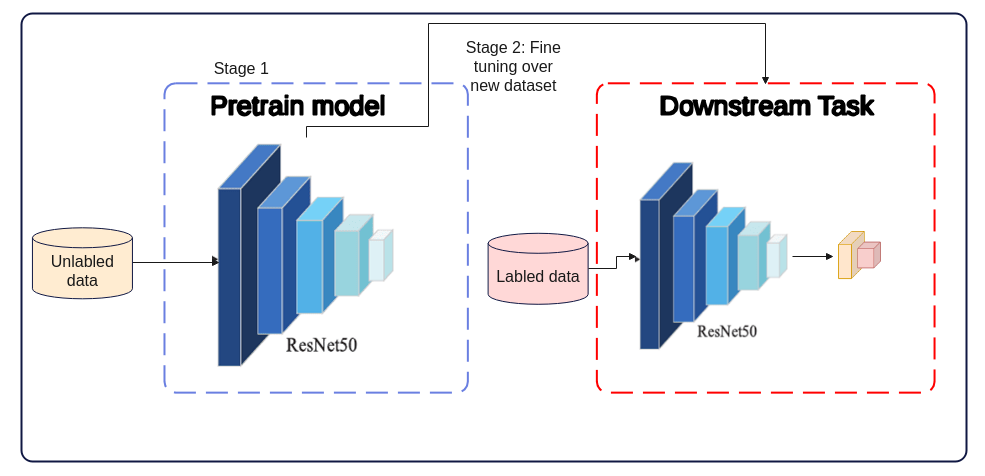}
\caption{General diagram of a self-supervised model trained over unlabeled data.} \label{fig1}
\end{figure}

is one of the emerging technologies in machine learning that eliminates the human interaction in labeling the data and uses unlabeled data directly in initializing a basic neural network by either creating pseudo labels or using positive or negative image pairs to learn representations \cite{albelwi2022survey}. Then the model is fine-tuned on a downstream task using a small labeled dataset. The general diagram in Fig. \ref{fig1} shows the general stages in self-supervised learning approach \cite{jaiswal2020survey}. Self-supervised learning is classified into two types: the contrastive and non-contrastive approaches (auxiliary pretext tasks) \cite{rani2023self}. The first approach is state-of-the-art and achieves higher performance in SSL, such as SimCLR \cite{chen2020simple}, SwAV \cite{caron2020unsupervised} and recently DINO \cite{oquab2023dinov2}. 
\squeezeup
\squeezeup
In general, the SimCLR framework consists of a data augmentation module, a base encoder, a projection head, and the contrastive loss function. The model produces a huge number of negative and positive samples, then the base encoder and head are trained to maximize the similarity between the augmented image. If those images are close to the original image (positive sample), then the contrastive loss produced will be low. Otherwise, the constructive loss increases.

\subsubsection{Downstream Task}
The most popular downstream tasks mentioned in almost all self-supervised approaches are either object detection, segmentation, or classification. Most self-supervised approaches are tested for such downstream tasks.

Researchers are highly interested in pose estimation models due to their ability to predict motion and ensure object safety, particularly for humans \cite{lan2022vision}. Numerous research studies have focused on detecting and creating skeletons for various subjects, including the entire human body \cite{zhang2020distribution}, facial gestures \cite{lin2021structure}, and hand movements \cite{jin2020whole}. Additionally, some research has explored skeleton generation for specific animals \cite{cao2019cross} and insects \cite{pereira2019fast}. Human pose estimation approaches can be categorized into two main types \cite{lan2022vision}: the top-down approach and the bottom-up approach. In the top-down approach, the model estimates the pose of individual objects in a single stage by leveraging object detection models to identify keypoints for each object independently. On the other hand, the bottom-up approach involves two stages: detecting keypoints for all objects in the scene using heatmaps and grouping the keypoints through associative embedding \cite{newell2017associative}  or graph neural network techniques \cite{jin2020differentiable}. The top-down approach has advantages over the bottom-up in terms of performance and accuracy, but it requires high computational cost, particularly in crowded and multi-object scenarios.

\section{Proposed Method}
The research on pose estimation for heavy and construction machines is limited, with most of the previous efforts focusing on supervised approaches that require a large labeled dataset. This paper aims to introduce a new method for pose estimation by utilizing state-of-the-art self-supervised learning and automatically generated simulator datasets to learn the main representation of objects in images and their features. The acquired knowledge can then be transferred to a real-time system using a small annotated dataset with only four main keypoints instead of the six keypoints mentioned in previous literature \cite{luo2020full}. Two points are removed as the model focuses on detecting the keypoints of the manipulator. Moreover, one keypoint for the bucket will be enough for the study of the movement of such machines as the geometry of the object will be considered.

\begin{figure}[h!]
\includegraphics[width=0.80\textwidth]{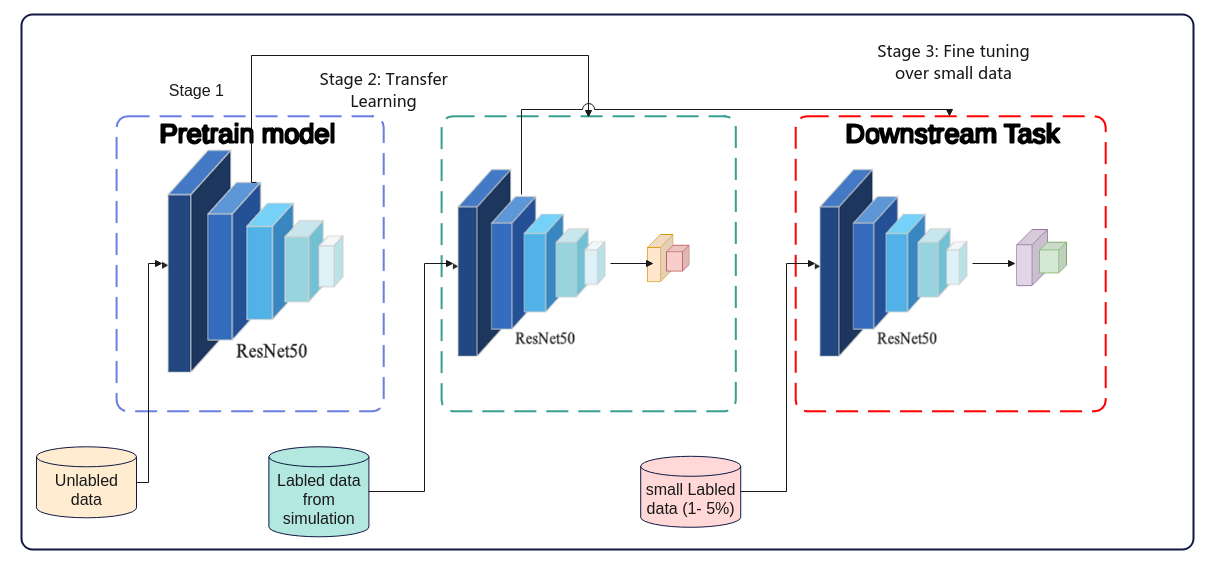}
\caption{Diagram of the approach with three-stage self-supervised paradigm.}\label{fig2}
\end{figure}

%The proposed model comprises three main stages: the self-supervised model serving as a pre-training stage with an unlabeled dataset, a feature extraction stage, and a fine-tuning stage with a small annotated dataset.
The proposed model consists of four main steps:
\begin{enumerate}[noitemsep]
    \item Generate synthetic images using simulators 
    \item Annotate those images automatically using a script based on the 4-keypoints definition in table\ref{tab:table1}. 
    \item Pre-train the deep neural network model based on the self-supervised approach with an unlabelled ACID dataset.
    \item Downstream task: 
    \subitem Knowledge transfer and feature extraction stage using a synthetically annotated dataset.
    \subitem Fine-tuning stage with a small annotated dataset from real images.
\end{enumerate}
%\squeezeup
Figure \ref{fig2} presents the general framework of the proposed approach. The latter two stages can be considered as high-level downstream tasks. The overall approach takes advantage of simulation environments such as Simscape Multibody and NVIDIA Isaac to generate and label data automatically. The usage of two different datasets from different simulators is necessary for higher variance.

%\squeezeup
In this research, a new direction of pose estimation and skeleton detection is presented by leveraging the strength of both the self-supervised learning approach and the dynamic simulation. The model learns the main features within the images using self-supervised learning without a need for annotation and then fine-tunes and downstreams using an annotated dataset generated automatically from the simulator. This approach reduces the human effort in labeling the dataset as well as increases the accuracy of common keypoint detectors which are based on the supervised learning paradigm. The proposed model is based on a top-down pose estimation approach which is easy to implement and can be developed using simulations.

\section{Experimental Setup}

\subsection{Datasets and Implementation}
As mentioned in the previous section, the proposed model consists of three main stages of knowledge transfer. Different types of datasets are used at each stage. At the pre-train self-supervised stage, the ACID dataset \cite{xiao2021development} of 2850 unannotated images was used to train the SimCLR approach. 

\begin{figure}[htp!]
    \centering
    \begin{subfigure}[b]{\textwidth}
        \centering
        \includegraphics[width=0.475\linewidth]{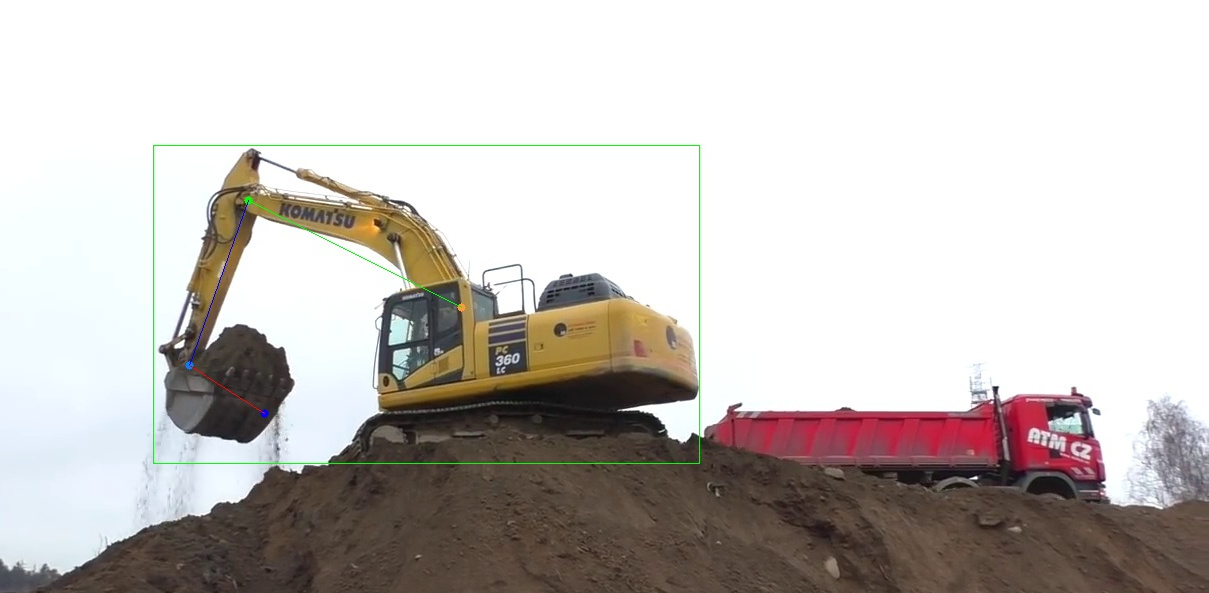}%
        \hfill
        \includegraphics[width=0.475\linewidth]{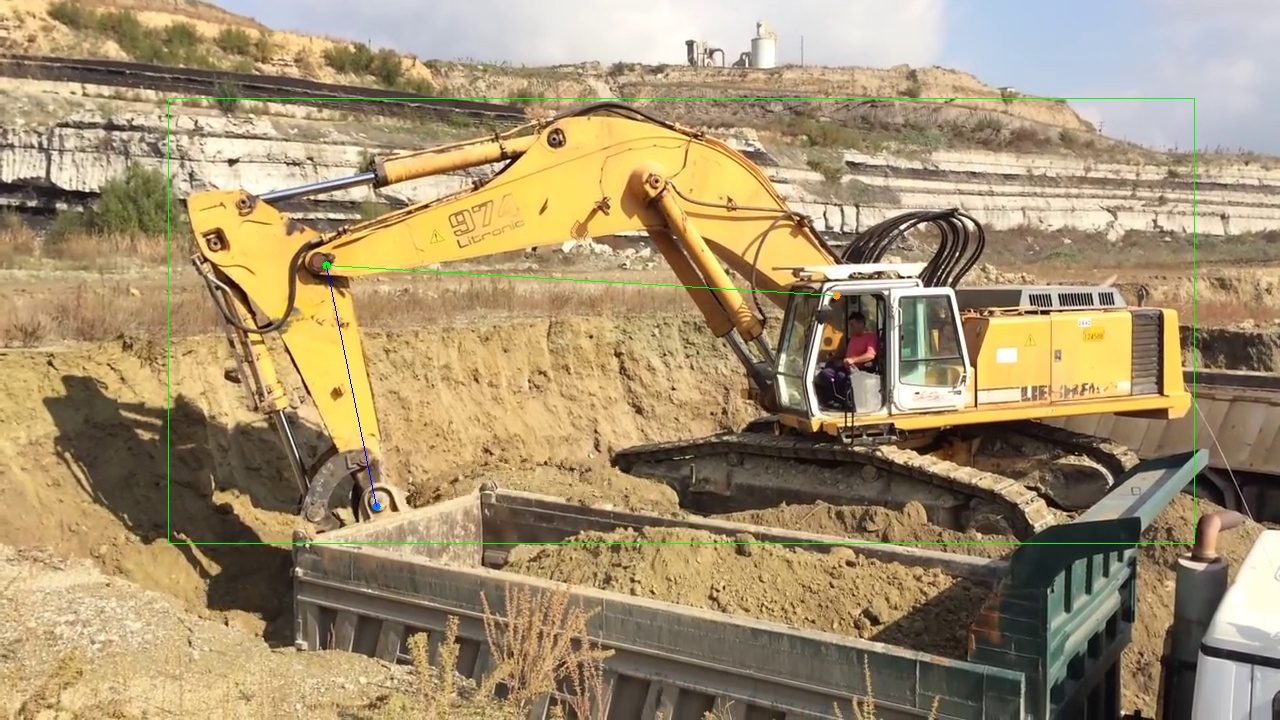}
    \end{subfigure}
    \caption{Image samples from ACID manually annotated for pose estimation.}\label{fig3:Samples from ACID pose}
\end{figure}

 On the other hand, the last two stages of downstream tasks require an annotated dataset with the basic keypoints of construction machines. 
To avoid the tedious and time-consuming task of annotating the ACID dataset, two additional simulated datasets have been generated and annotated automatically using the NVIDIA Isaac (500 images) and MATLAB Simscape Multibody (300 images) simulators, which are 17.5\% and 10.5\% of the original ACID dataset, respectively. Those datasets are used later in the feature extraction stage for the pose estimator model based on both backbones of ResNet-50 and HRNet-W32. 
 Around 3.5\% of the whole ACID dataset has been manually annotated with the basic four keypoints mentioned in the right column of Table \ref{tab:table1}. This dataset is used in the last stage to fine-tune the model on a downstream task of pose estimation and map the features between simulated and real situations. As mentioned in the related work section, another dataset \cite{luo2020full} for real excavators consisting of 6405 images with six keypoints has been used in feature extraction to test its effect on the performance of the model. Figure \ref{fig3:Samples from ACID pose} shows samples from part of the ACID dataset that has been annotated manually in the pose estimation. All training was done with an NVIDIA Tesla V100 with 64GB memory. The self-supervised models for both backbones are trained with a batch size of 256.

%multi images for acid original and manually annotated with pose 
\squeezeup
\subsection{Evaluation Metrics}
The effectiveness of the proposed approach has been evaluated using the two main metrics Average Precision (AP) and Average Recall (AR) with different Object Keypoint Similarity (OKS) thresholds: OKS=0.50:0.05:0.95, OKS=0.50, OKS=0.75, and different object sizes M (medium) and L (large). The OKS is measured using

\begin{equation}
  \mathrm{OKS} = \sum_{i} [\mathrm{exp}(-d_{i}^2/2s^2k_{i}^2)\delta (\upsilon_{i}> 0)]/\sum_{i} [\delta (\upsilon_{i} > 0)]
\end{equation}

where
$d_{i}$ is the Euclidean distance between the ground truth and predicted keypoint, $s$ is the object segment area, $k$ is the per-keypoint constant that controls fall-off, and $v_{i}$ is the visibility flag that can be 0, 1, or 2 for not labeled, labeled but not visible, and visible and labeled, respectively. The OKS is within 0 and 1, where 1 represents perfect predictions while 0 represents poor ones.  

\section{Results and Discussion}
\subsection{Main Results}
In this subsection, two scenarios of the self-supervised feature extraction approach are presented. The first scenario involves employing a SimCLR model trained on a standard ImageNet dataset using a traditional constructive learning approach. In this case, the model is directly utilized for pose estimation downstream tasks using the small annotated ACID dataset. Table \ref{tab2} shows the result of such a scenario in comparison to the supervised learning approach for models based on ResNet-50 and HRNet-w32 backbones.

%###### beginning of table 2 ##################%
\begin{table}[h]
\caption{The first scenario: supervised learning versus normal self-supervised learning approach for learning features for a pose estimation downstream task. There is no intermediate stage in this approach.}\label{tab2}
\resizebox{\textwidth}{!}{
\centering
%\begin{adjustbox}{max width=\textwidth}
\begin{tabular}{|l|l|l|c|c|c|c|c|c|c|c|c|c|l|}
\hline
Model & Datasets &Status &AP &AP$^{50}$ & AP$^{75}$&AP$^{M}$	&AP$^{L}$& AR& AR$^{50}$& AR$^{75}$& AR$^{M}$ &AR$^{L}$&Epochs\\
\hline
\multirow{2}{*}{ResNet-50}&ACID$_p$& Sup & \textbf{0.4455}&\textbf{1.0} & \textbf{0.3307}&	-1.0& \textbf{0.4455}&\textbf{0.4867} &\textbf{1.0}& \textbf{0.4667}&	-1.0&\textbf{0.4867}& 120\\
&ImageNet& Self-Sup &0.37962 & 0.80528& 0.17327&	-1.0& 0.37962&0.42  &0.86667& 0.2&	-1.0&0.42 & 590\\
\hline
\multirow{2}{*}{HRNet-W32}&ACID$_p$ & Sup& \textbf{0.4228} & \textbf{1.0}& \textbf{0.3187}& -1.0&  \textbf{0.4228}&\textbf{0.520} &\textbf{1.0}& \textbf{0.5333}&	-1.0& \textbf{0.520}& 120\\	
&ImageNet & Self-Sup& 0.3832 & 0.7958& 0.2911&-1.0&0.3832 &0.4200 &0.8667&0.3333 &-1.0	& 0.4200& 450\\	
\hline
\end{tabular}
%\end{adjustbox}
}
\end{table}
%###### end table ##################%

On the other hand, the second scenario uses the proposed approach with three stages of self-supervised feature extraction, transfer learning, and fine-tuning to get more accurate pose estimation results for construction machines. Table \ref{tab3} shows the results of this proposed self-supervised three-stages-fine-tuning approach. The performance for both scenarios is given in terms of Average Precision (AP), Average Recall (AR), and number of epochs to reach the optimal weights. 

%###### beginning of table 3 ##################% 
\begin{table}[h]
\caption{The second scenario : results of fine tuning ResNet-50 on small ACID dataset (3.5\% of whole ACID dataset) from different transfer learning models used one of three main pose estimation datasets. The initial self-supervised feature extraction was trained over the whole ACID dataset.}\label{tab3}
%\begin{center}
\resizebox{\textwidth}{!}{
\begin{tabular}{|l|l|l|c|c|c|c|c|c|c|c|c|c|l|}
\hline
 Model & Data &Status &AP &AP$^{50}$ & AP$^{75}$&AP$^{M}$	&AP$^{L}$& AR& AR$^{50}$& AR$^{75}$& AR$^{M}$ &AR$^{L}$&Epochs \\
\hline
\multirow{3}{*}{ResNet-50}&6 keys& Self-Sup&\textbf{ 0.4913}& 0.9261 &	\textbf{0.3964}& -1.0&	\textbf{0.4913}&	\textbf{0.5400}&	0.9333&	\textbf{0.5333}&	-1.0&	\textbf{0.5400}	&\textit{\underline{100}}\\
&Nvidia & Self-Sup& 0.405 &	0.861& 0.352&	-1.0	&	0.405&	0.440	&	0.867	&0.467&	-1.0&	0.440& 560\\	
&Matlab & Self-Sup&0.4562 &1.0	 & 0.3030 & -1.0 &	0.4562&  0.5133&	1.0&0.4000 & -1.0& 0.5133 &	440 \\	
\hline
\multirow{3}{*}{HRNet-W32}&6 keys& Self-Sup& \textbf{0.5145	}&\textbf{1.0	}&\textbf{0.3634	}&-1.0	&\textbf{0.5145	}&\textbf{0.5667	}&\textbf{1.0	}&\textbf{0.4667	}&-1.0	&\textbf{0.5667}&\textit{\underline{330}}\\
&Nvidia & Self-Sup& 0.2858	&0.61174	&0.10396	&-1.0	&0.2858	&0.32667	&0.32667	&0.13333	&-1.0	&0.32667&510\\	
&Matlab & Self-Sup&	0.1477	&0.4738	&0.0901	&-1.0	&0.1477	&0.26	&0.6667	&0.20	&-1.0	&0.260	  &480\\	
\hline
\end{tabular}}
%\end{center}
\end{table}
%###### end ##################%

\begin{figure}[!ht]
\subfloat[Self-supervised model based on ResNet-50]{%
  \includegraphics[clip,width=0.85\columnwidth]{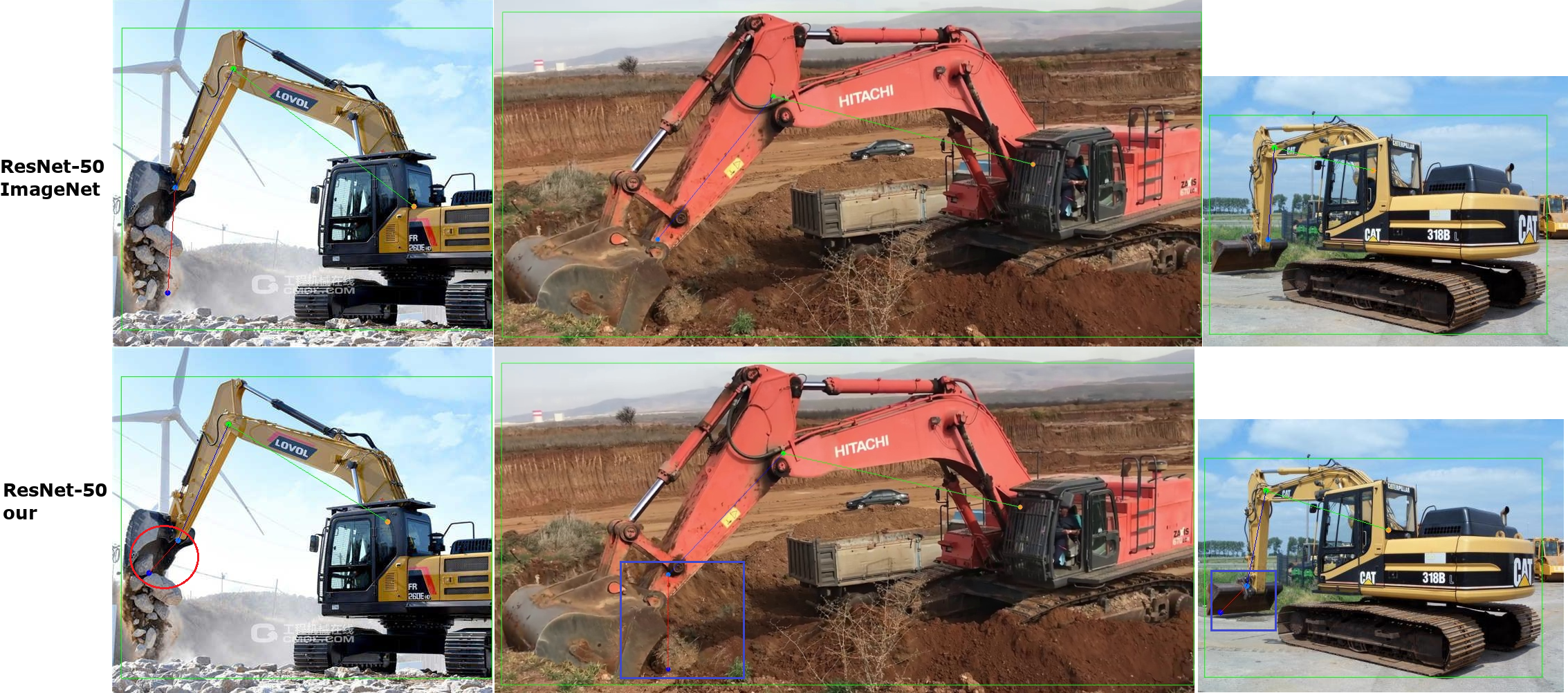}%
}

\subfloat[Self-supervised model based on HRNet-W32]{%
  \includegraphics[clip,width=0.85\columnwidth]{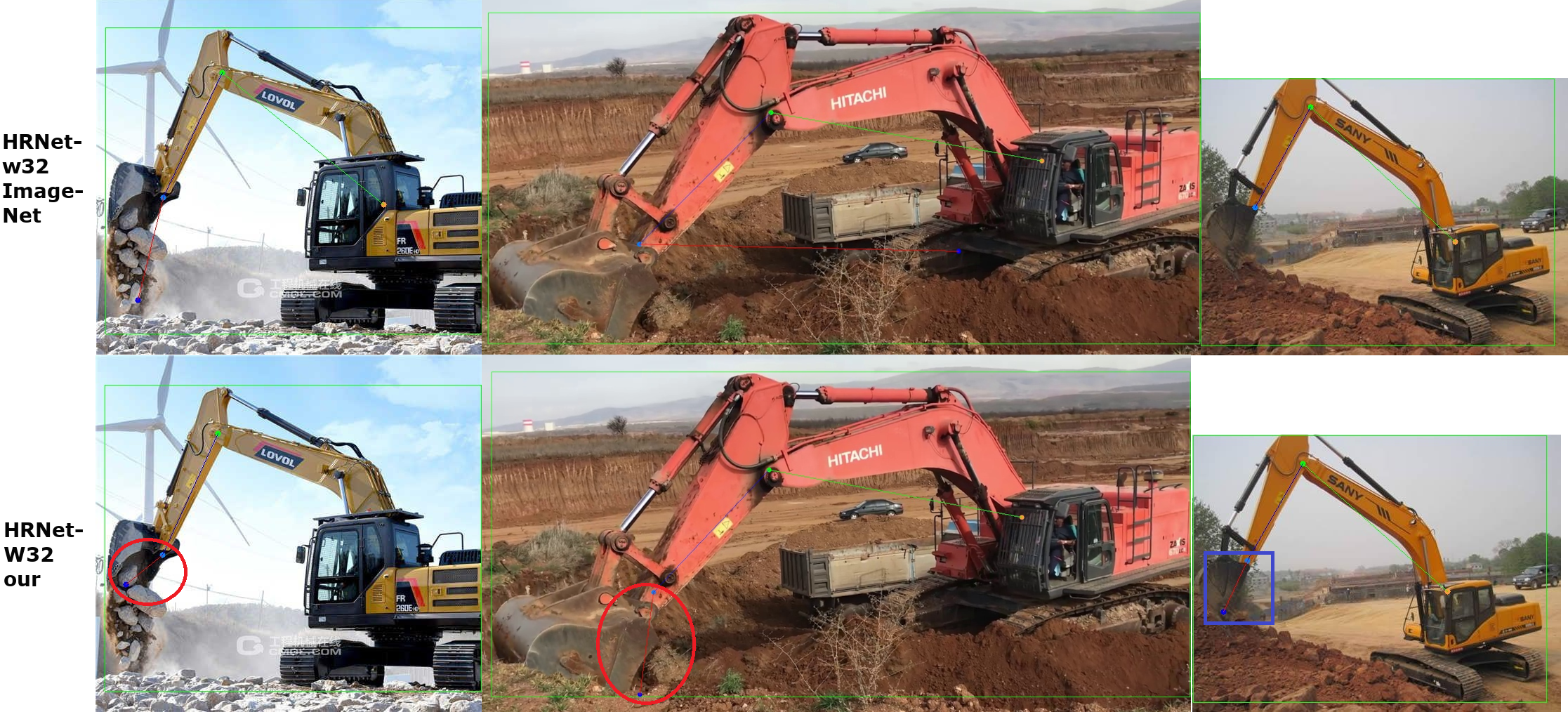}%
}
\caption{Samples results of the self-supervised model originally pretrained with ImageNet, and the proposed approach. Red circle shows a new keypoint detection, while the blue box shows a correction in the detection. }\label{fig5:results as image}
\end{figure}

\begin{figure}[!ht]
     \centering
     \begin{subfigure}[b]{0.45\textwidth}
         \centering
         \includegraphics[width=\textwidth]{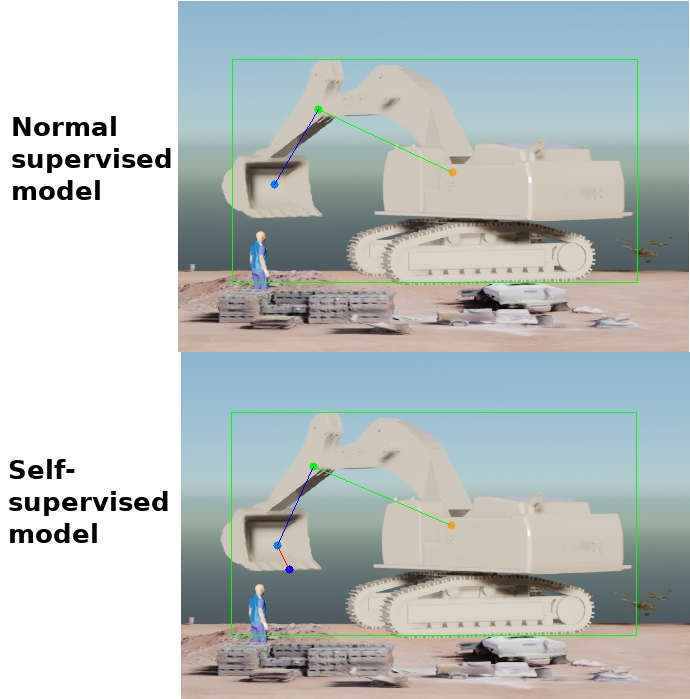}
         \caption{Samples from NVIDIA simulator.}
         \label{fig:y equals x}
     \end{subfigure}
     \hfill
     \begin{subfigure}[b]{0.4\textwidth}
         \centering
         \includegraphics[width=\textwidth]{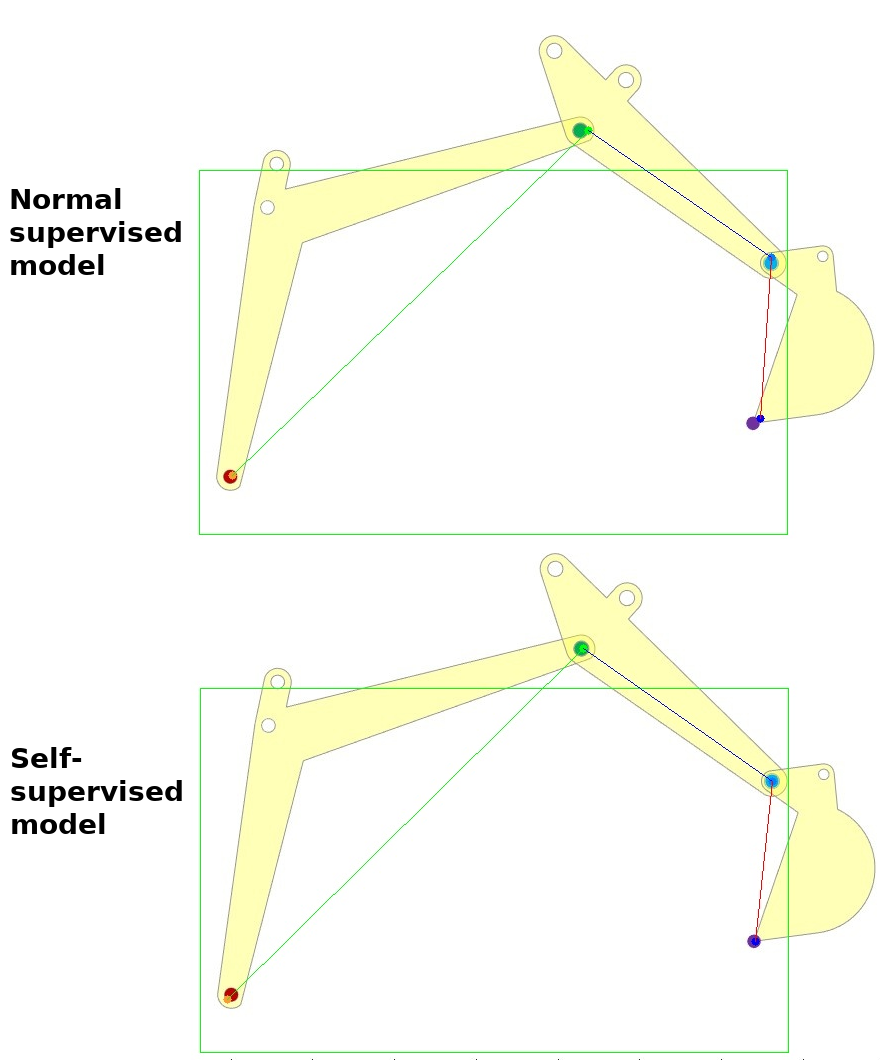}
         \caption{Samples from MATLAB simulator.}
         \label{fig:three sin x}
     \end{subfigure}
        \caption{Samples from supervised and self-supervised models trained on datasets from NVIDIA Isaac and Simscape Multibody using the ResNet-50 backbone.}
        \label{fig6:results_from_simulation}
\end{figure}

The self-supervised stage is trained over the ACID whole dataset. Then the learning of downstream task is transferred using three different datasets (two simulated datasets from NVIDIA Issac and MATLAB as well as one real dataset with 6 keypoints ) and finally fine-tuned on the small annotated ACID dataset with only 4 keypoints. In addition, the results of the middle stage with a simulated dataset from NVIDIA Isaac and MATLAB Simscape Multibody are mentioned in Table \ref{tab4}. The self-supervised approach is compared to the supervised approach in terms of Average Precision (AP), Average Recall (AR), and optimal epoch for models based on both ResNet-50 and HRNet-W32 backbones.
%###### beginning of table 4 ##################%
% \usepackage{multirow}
\begin{table}[h]
\caption{Results of pose estimation model training with both supervised and self-supervised approaches on the two simulated datasets with different backbones. Feature extraction in self-supervised stage was trained over whole ACID dataset.}\label{tab4}
%\begin{center}
\resizebox{0.70\textwidth}{!}{
\begin{tabular}{|lllllllll|}
\hline
\multicolumn{1}{|l|}{Status} & \multicolumn{1}{l|}{Model}                      & \multicolumn{1}{l|}{AP}             & \multicolumn{1}{l|}{AR}             & \multicolumn{1}{l|}{Epochs} & \multicolumn{1}{l|}{Model}                      & \multicolumn{1}{l|}{AP}             & \multicolumn{1}{l|}{AR}             & Epochs \\ \hline
\multicolumn{9}{|l|}{Dataset: NVIDIA simulated dataset}                                                                                                                                                        \\ \hline
\multicolumn{1}{|l|}{Supervised}    & \multicolumn{1}{l|}{\multirow{2}{*}{ResNet-50}} & \multicolumn{1}{l|}{0.817}          & \multicolumn{1}{l|}{0.863}          & \multicolumn{1}{l|}{150}   & \multicolumn{1}{l|}{\multirow{2}{*}{HRNet-w32}} & \multicolumn{1}{l|}{0.814}          & \multicolumn{1}{l|}{0.871}          & 380   \\ \cline{1-1} \cline{3-5} \cline{7-9} 
\multicolumn{1}{|l|}{Self-sup}   & \multicolumn{1}{l|}{}                           & \multicolumn{1}{l|}{\textbf{0.834}} & \multicolumn{1}{l|}{\textbf{0.888}} & \multicolumn{1}{l|}{200}   & \multicolumn{1}{l|}{}                           & \multicolumn{1}{l|}{\textbf{0.940}} & \multicolumn{1}{l|}{\textbf{0.961}} & 600   \\ \hline
\multicolumn{9}{|l|}{Dataset: MATLAB simulated dataset}                                                                                \\ \hline
\multicolumn{1}{|l|}{Supervised}    & \multicolumn{1}{l|}{\multirow{2}{*}{ResNet-50}} & \multicolumn{1}{l|}{0.688}          & \multicolumn{1}{l|}{0.696}          & \multicolumn{1}{l|}{120}   & \multicolumn{1}{l|}{\multirow{2}{*}{HRNet-w32}} & \multicolumn{1}{l|}{\textbf{0.690}} & \multicolumn{1}{l|}{0.696}          & 90    \\ \cline{1-1} \cline{3-5} \cline{7-9} 
\multicolumn{1}{|l|}{Self-sup}   & \multicolumn{1}{l|}{}                           & \multicolumn{1}{l|}{\textbf{0.696}} & \multicolumn{1}{l|}{\textbf{0.699}} & \multicolumn{1}{l|}{130}   & \multicolumn{1}{l|}{}                           & \multicolumn{1}{l|}{0.684}          & \multicolumn{1}{l|}{\textbf{0.7}}   & 550   \\ 
 \hline
\end{tabular}}
\end{table}
%###### end  ##################%
%
For a comprehensive presentation of the data, we provide images samples trained using conventional supervision and our novel self-supervised model, applied to authentic images of the excavator. The visual representations of both approaches are displayed in Fig. \ref{fig5:results as image}, additional results from simulations are given in Fig. \ref{fig6:results_from_simulation}.
%figure from real images 
\squeezeup

% figure from simulation 

\subsection{Discussion}
The proposed model of a self-supervised approach trained over three stages boosts up the performance of the pose estimation task by 11\% and 13\% in comparison to the SimCLR self-supervised baseline for the backbones of ResNet and HR-Net, respectively. In addition, it reduces the number of epochs needed during the training for both encoder models which are the basic deep neural structure in the most recent pose estimation models. It is noticeable that this model outperforms the supervised approach for pose estimation for small data by 4.6\% and 9.2\% for the mentioned backbones,  respectively, after transfer learning on a 6 keypoints real dataset.
\squeezeup
\begin{figure}[!htp]
\includegraphics[width=0.75\textwidth]{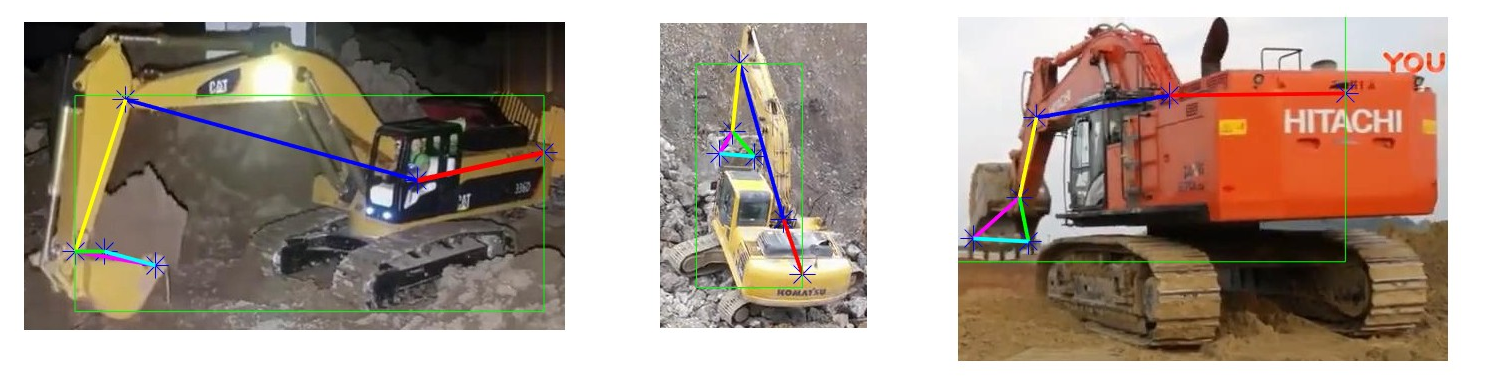}
\caption{Samples of automatic bounding box computation in green labeling using the minimum bounding box of the six keypoints of excavator pose.} \label{fig4:uncorrect_bbox_anno}
\end{figure}

The results in tables \ref{tab3} and \ref{tab4} show that the accuracy from training the proposed three-stage model with such simulation datasets of NVIDIA Issac and MATLAB is not as higher as expected. One of the reasons behind that is that the downstream task depends on having some annotated dataset. In this proposed approach, usage of the top-down pose estimation as a downstream task requires accurately labeled data, especially in terms of bounding boxes. Since both the Matlab and six-keypoints datasets do not have any boxes over the desired object, an approximate method called Minimum bounding box \cite{chan2001determination} is used to create such boxes without using human label effort. This method is not so accurate due to multiple reasons: first, we try to use few points to create such boxes, second, those points are the main keypoints that can not present the exact bounding of the excavator in some cases, e.g.~when the picture of the machine is taken from the back view. Figure \ref{fig4:uncorrect_bbox_anno} shows such cases. Thus an accurate automatic labeling of the data from the simulation is of high importance to enhance the accuracy of such a top-down pose estimator which will be addressed in future work.
\section{Conclusion and Future Work}
In this paper, a new pose estimation approach is proposed inherited from both the self-supervised SimCLR model and dynamic simulation. The self-supervised model is used to teach the pose estimator the basic features that help in detecting the keypoints of construction machines, while the dynamic simulators are used to generate a synthetic dataset with annotations to enhance the downstream tasks. Implementing such a synthetic dataset within a self-supervised approach is an emerging area of research, which has been addressed within this work. 

With around 3.5\% annotated data of real images from the overall ACID dataset, the proposed model was sufficient to detect the main keypoints of construction machines. It increases the accuracy of pose estimation for construction machines for both backbones ResNet-50 and HRNet-W32 by 11\% and 13\%, respectively compared to pre-train the model using the ImageNet dataset in one stage. Moreover, the pretraining using a self-supervised approach over the ACID dataset and fine-tuning with multiple stages over a smaller dataset helps the model to reach the optimal weights faster with less number of epochs. This approach is boosting the usage of real-time dynamic simulators in improved computer vision models in the agriculture and construction fields that have many limitations in gathering and labeling huge amounts of datasets.

In future work, we plan to investigate and design a pose estimation mode using the recent state-of-the-art SSL Dino v2 model. Using domain adaptation approaches can be tested and integrated with our model to enhance the pose estimation for other types of construction machines as well as autonomous driving machines. In addition, we will integrate our deep neural model into a more complex approach for motion prediction to ensure safety in construction fields. 
%\subsubsection{Acknowledgements} .
\bibliographystyle{splncs04}
%\typeout{}
\bibliography{egbib}
%\end{thebibliography}
\end{document}